\documentclass[a4paper,11pt]{article}

\usepackage{geometry}
\usepackage{titling}
\usepackage[utf8]{inputenc}
\DeclareUnicodeCharacter{2212}{-}
\usepackage{amsmath}
\usepackage{microtype}
\usepackage{siunitx}
\usepackage{lineno}
\usepackage{authblk}
\usepackage{lscape}
\usepackage{graphicx}
\usepackage{subcaption}
\usepackage[font={footnotesize},labelfont=bf]{caption}
\usepackage[super,numbers,sort&compress]{natbib}
\usepackage{booktabs}
\usepackage{blindtext}
\usepackage{booktabs,multirow,array}
\usepackage{tikz}
\usepackage{verbatim}
\usepackage{chemfig}
\usepackage{gensymb}
\usepackage{textcomp}
\usepackage{setspace}
\usepackage{multicol}
\usepackage{pgfplots,pgfplotstable}
\pgfplotsset{compat=1.17}
\usepackage{afterpage}
\usepackage{xtab,afterpage}
\usepackage{amssymb}
\usepackage{rotating}
\usepackage{scrextend}
\usepackage{tcolorbox}
\usepackage{xr}
\usepackage[hidelinks]{hyperref}
\hypersetup{
    colorlinks=true,
    linkcolor=blue,
    filecolor=magenta,      
    urlcolor=cyan,
    }

\usepackage{cleveref}
\usepackage{algorithm}
\usepackage[noend]{algpseudocode}
\usepackage{enumitem}
\setlist[itemize]{leftmargin=*}
\usepackage{microtype}
\usepackage{float}

\algnewcommand\algorithmicinput{\textbf{Preprocessing:}}
\algnewcommand\Preprocessing{\item[\algorithmicinput]}

\usepackage{arxiv}

\usepackage{txfonts}
\usepackage{soul}

\begin{document}

\title{SAMBA: A Trainable Segmentation Web-App with Smart Labelling}

\author[1]{{Ronan Docherty}
\thanks{ronan.docherty18@imperial.ac.uk}\ \ }

\author[2]{{Isaac Squires}}

\author[2]{{Antonis Vamvakeros}}

\author[2]{Samuel J. Cooper \thanks{samuel.cooper@imperial.ac.uk}\ \ }

\affil[1]{{\textit{\footnotesize Department of Materials, Imperial College London, London SW7 2DB}}}

\affil[2]{{\textit{\footnotesize Dyson School of Design Engineering, Imperial College London, London SW7 2DB}}}

\lhead{\scshape Docherty \textit{et al.}}
\chead{\scshape Trainable Segmentation Web-App}
\rhead{\scshape Preprint}

\maketitle

\begin{abstract}
\begin{center}
\begin{minipage}{0.85\textwidth}
{\small Segmentation is the assigning of a semantic class to every pixel in an image and is a prerequisite for various statistical analysis tasks in materials science, like phase quantification, physics simulations or morphological characterization. The wide range of length scales, imaging techniques and materials studied in materials science means any segmentation algorithm must generalise to unseen data and support abstract, user-defined semantic classes. Trainable segmentation is a popular interactive segmentation paradigm where a classifier is trained to map from image features to user drawn labels. SAMBA is a trainable segmentation tool that uses Meta's Segment Anything Model (SAM) for fast, high-quality label suggestions and a random forest classifier for robust, generalizable segmentations. It is accessible in the browser (\href{https://www.sambasegment.com/}{https://www.sambasegment.com/}) without the need to download any external dependencies. The segmentation backend is run in the cloud, so does not require the user to have powerful hardware.
}

\end{minipage}
\end{center}
\end{abstract}
\vspace{.2cm}
\begin{multicols}{2}

\section{Statement of Need}
\label{intro}

Trainable segmentation tools like ilastik \cite{ilastik} and Trainable Weka Segmentation \cite{weka} are popular options for image segmentation in biology and materials science due to their flexibility. Both apply common image processing operations like blurs, edge detection, and texture filters to create a feature stack, then train a classifier (usually a random forest ensemble) to map from the stack to user drawn labels on a pixelwise basis. ilastix offers fast segmentations and a responsive live-view, whereas Trainable Weka Segmentation is part of and interoperates with the FIJI ImageJ library \cite{FIJI}.

Accurate, representative labels are important for segmentation quality, but tooling is limited to polygons and/or pixel brushes which can make labelling boundaries time-consuming. SAMBA (Segment Anything Model Based App) uses the recent object-detecting foundation model Segment Anything Model \cite{SAM} as a `Smart Labelling' tool, which offers suggestions for associated objects as the cursor is moved which can then be added as labels. This `Smart Labelling' can quickly pick out portions of the same phase and is resilient against noise, varying exposure, and certain types of artefacts \textit{etc.} SAM has been applied to cell microscopy annotation and segmentation \cite{micro-SAM}, but not to multi-phase segmentation.

SAMBA is also unique in being fully web-based, meaning it can be used without downloading a large binary, installing libraries or requiring significant local computational power. An associated gallery page allows users to contribute to and view an open dataset of micrographs, labels, segmentations and experimental metadata. It is designed to be usable by researchers regardless of prior programming or image analysis experience.

\begin{figure*}
\centering
    \includegraphics[width=0.9\linewidth]{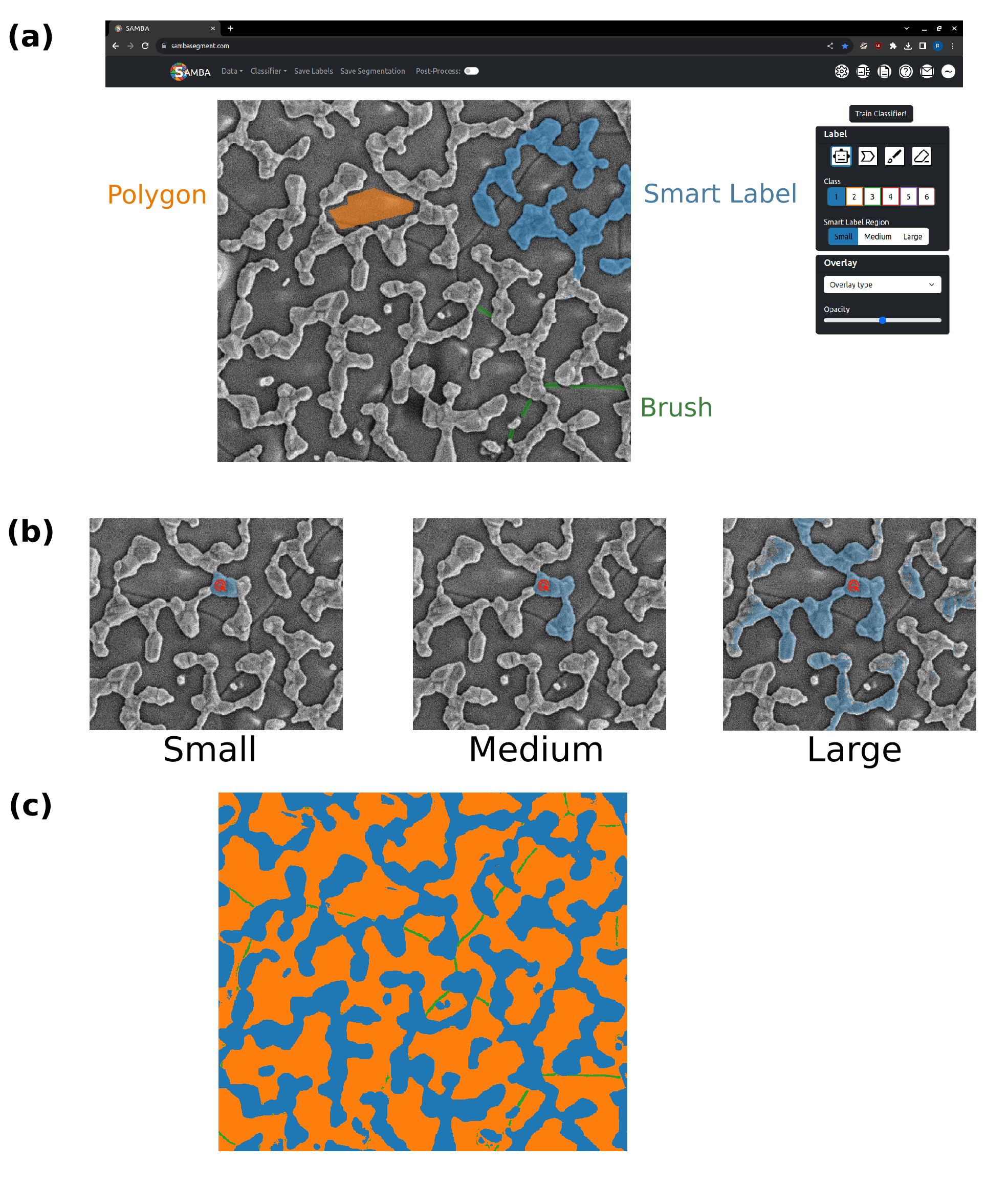}
    \caption{\textbf{(a)} a screenshot of the GUI of SAMBA, highlighting the different possible labelling types. \textbf{(b)} shows the impact of changing the `Smart Labelling' scale parameter for the same cursor location, enabling fine-grained segmentation of small objects or rapid segmentation of large phases. \textbf{(c)} the resulting multi-phase segmentation from the random forest backend.  }
    \label{fig:gui}
\end{figure*}

\section{Design and Usage}
\label{sec:design} 

SAM is a \textit{promptable}, open-source macroscopic object segmentation model, trained on 1 billion masks over 11 million images. It feeds a prompt (click, bounding box, mask or text) alongside an embedding of an image generated by a pretrained autoencoder \cite{ViT-22B} through a lightweight decoder network to produce a segmentation for the prompt (\textit{i.e,} the object at the mouse cursor). The decoder is lightweight enough that it can be run in the browser in real-time when exported via the ONNX framework \cite{ONNX}, though the embedding must still be generated by a large autoencoder model running on a computer with PyTorch. Despite its impressive zero-shot performance, SAM is primarily an object-detection model and cannot be used for the semantic phase segmentation desired in materials science. However, its understanding of shape, texture and other lower-level features apply well to instances of certain phases.

This provides the motivation for and structure of SAMBA: a Python web server supplies the image embedding (and later performs random forest segmentation) for the user's image, then browser-based labelling is performed by client-side object segmentations as label suggestions. Labels are confirmed with a left click, and a right click switches length scales of the smart labelling - this was achieved by exposing SAM's hidden multi-mask output. Normally, SAM produces three masks per prompt - each representing a different length scale and with an associated internal quality estimate. Giving the user the choice of length scales allows greater flexibility, which is useful given SAM is not designed for micrographs, and therefore its internal estimate of quality may not always match the user's.

Standard drawing tools like a variable width brush, polygon, and eraser allow the user to create labels when SAM struggles, for example with very high frequency spatial features. Previous labels are not overwritten (unless erased), so new labels can be added on top for boundary fine-tuning or rapid whole-image labelling. These labels can then be downloaded for use in offline workflows, like training task-specific neural networks.

After labelling is complete, the Python web server trains a random forest ensemble to map the feature stack (computed in the background) to the labels. The feature stack creation involves a Python reimplementation of most of the image processing operations outlined in Trainable Weka Segmentation. This is achieved predominantly using the scikit-image library \cite{scikit-image}. Once completed, the resulting segmentation of the full image is returned to the user as a variable opacity overlay, allowing for additional labels to refine the output. An optional overlay highlights pixels the classifier is least certain about, which could be useful starting points for new user labels. Improvements in this `Human-In-The-Loop' (HITL) feedback (\textit{i.e,} feature-weighted uncertainty highlights) are interesting avenues for future work. The trained classifier is a scikit-learn object \cite{scikit-learn} - it can be downloaded and applied to future data either on SAMBA or in other Python workflows.

SAMBA has a clean, responsive user interface implemented in React and Typescript. It works for a range of common image file types (.png, .jpeg, .tiff), as well as large .tiff images and .tiff stacks. Local hosting is simple and advisable if the user is handling sensitive data or wishes to benefit from GPU acceleration. A desktop version that supports larger files, 3D segmentation and has deeper PyTorch integration is planned. SAMBA has an associated gallery, which users can optionally publish to once a segmentation is complete. It is hoped this will provide the foundation for a varied, open source library of high-quality image data and encourage further development of generalist machine learning models in materials science. A user manual, instructions for contributing and setup instructions for local hosting is available in the \href{https://github.com/tldr-group/samba-web}{public repository}.

\section*{Code Availability}
The code needed to host a local version of the web-app is available at \url{https://github.com/tldr-group/samba-web} with an MIT license agreement.

\section*{Acknowledgements}

This work was supported by funding from the EPRSC and SFI Centre for Doctoral Training in Advanced Characterisation of Materials (EP/S023259/1 recieved by RD) and the Royal Society (IF\textbackslash R2\textbackslash 222059 received by AV as a Royal Society Industry Fellow).

The authors would like to thank other members of the TLDR group for their testing and feedback.


\section*{Competing interests}
The authors declare no competing interests.

\section*{References}
\addcontentsline{toc}{section}{References}

\def\addvspace#1{}

	\renewcommand{\refname}{ \vspace{-\baselineskip}\vspace{-1.1mm} }
	\bibliographystyle{abbrv}
    \bibliography{main}

\end{multicols}

\end{document}